%% file: Main.tex
\documentclass[lettersize,journal]{IEEEtran}
\usepackage{amsmath,amsfonts}
\usepackage{algorithmic}
\usepackage{array}
\usepackage[caption=false,font=normalsize,labelfont=sf,textfont=sf]{subfig}
\usepackage{textcomp}
\usepackage{stfloats}
\usepackage{url}
\usepackage{verbatim}
\usepackage{graphicx}
\def\BibTeX{{\rm B\kern-.05em{\sc i\kern-.025em b}\kern-.08em
    T\kern-.1667em\lower.7ex\hbox{E}\kern-.125emX}}
\usepackage{balance}

\usepackage[linesnumbered, ruled, noend]{algorithm2e}
\SetEndCharOfAlgoLine{}
\SetKwInput{KwData}{The data}
\SetKwInput{KwResult}{The result}
\SetKwData{KwInput}{Input}
\SetKwData{KwOutput}{Output}

\usepackage[switch]{lineno}
\usepackage[table, xcdraw]{xcolor}
\usepackage{multirow}
\usepackage{booktabs}
\usepackage{cite}

\title{\LARGE \bf
HuBE: Cross-Embodiment \textit{Hu}man-like \textit{B}ehavior \\ \textit{E}xecution for Humanoid Robots
}

\author{Shipeng Lyu$^{\dag}$, Fangyuan Wang$^{\dag}$, Weiwei Lin, Luhao Zhu, \\
David Navarro-Alarcon$^{*}$ and Guodong Guo
\thanks{This work is supported in part by the Research Grants Council (RGC) of Hong Kong under grant 15231023, and in part by the PolyU-EIT Collaborative PhD Training Programme under application number 220766263. \textit{Corresponding author: David Navarro-Alarcon.}}
\thanks{S. Lyu, F. Wang, and D. Navarro-Alarcon are with the Department of Mechanical Engineering, The Hong Kong Polytechnic University (PolyU), Kowloon, Hong Kong. D. Navarro-Alarcon is also with the Research Institute for Smart Ageing (RISA), PolyU. (e-mail: shipeng.lyu, fangyuan.wang@connect.polyu.hk, dnavar@polyu.edu.hk)}%
\thanks{S. Lyu, F. Wang, W. Lin, L. Zhu and G. Guo are with the Ningbo Institute of Digital Twin, Eastern Institute of Technology (EIT), China. (e-mail: gdguo@eitech.edu.cn)}%
\thanks{$\dag$ Equal contribution; * Corresponding author}
}

\begin{document}

\maketitle
\thispagestyle{empty}
\pagestyle{empty}

\begin{abstract}
Achieving both behavioral similarity and appropriateness in human-like motion generation for humanoid robot remains an open challenge, further compounded by the lack of cross-embodiment adaptability.
To address this problem, we propose \textit{HuBE}, a bi-level closed-loop framework that integrates robot state, goal poses, and contextual situations to generate human-like behaviors, ensuring both behavioral similarity and appropriateness, and eliminating structural mismatches between motion generation and execution.
To support this framework, we construct HPose, a context-enriched dataset featuring fine-grained situational annotations.
Furthermore, we introduce a bone scaling-based data augmentation strategy that ensures millimeter-level compatibility across heterogeneous humanoid robots. 
Comprehensive evaluations on multiple commercial platforms demonstrate that \textit{HuBE} significantly improves motion similarity, behavioral appropriateness, and computational efficiency over state-of-the-art baselines, establishing a solid foundation for transferable and human-like behavior execution across diverse humanoid robots.  
\end{abstract}

\input{Section/Introduction}
\input{Section/Related_work}

\input{Section/Methodology}

\input{Section/Experiment}
\input{Section/Conclusion}


 \bibliographystyle{IEEEtran}
 \bibliography{IEEEabrv,References.bib}

\end{document}

%% file: Section/Introduction.tex
\section{Introduction}
\label{sec:intro}
\begin{figure}[t]
    \centering
    \includegraphics[width=\linewidth]{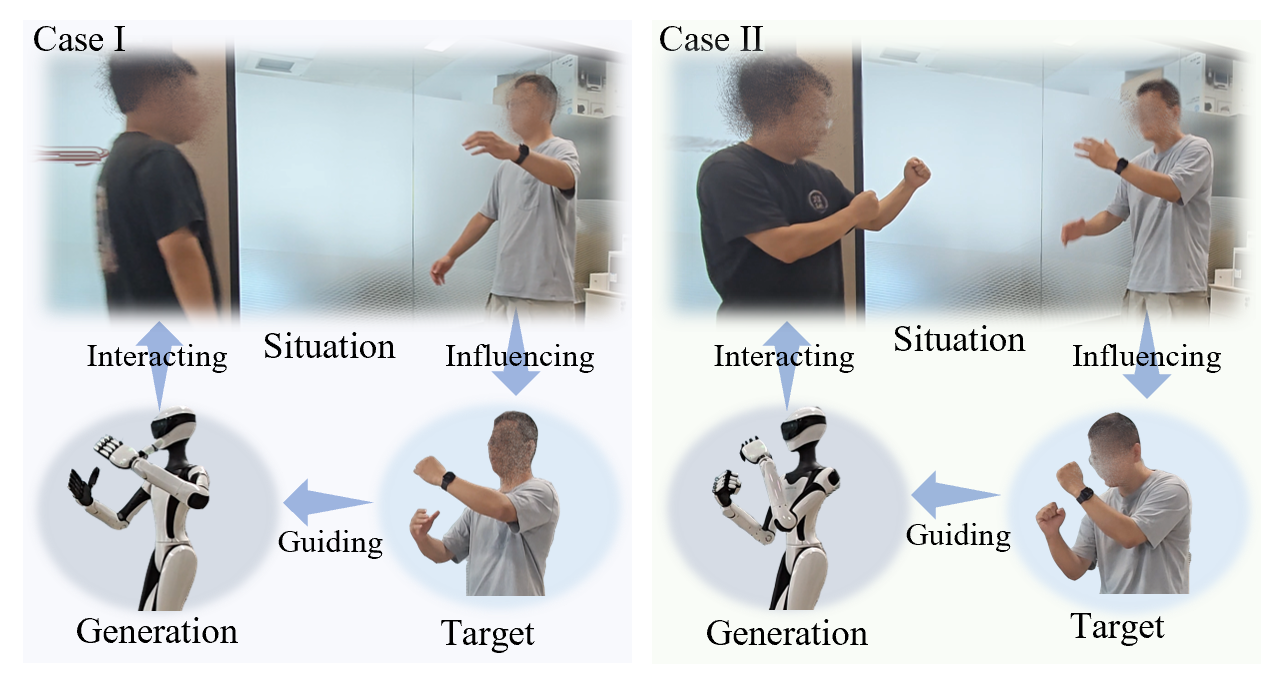}
    \caption{\textbf{Contextual semantics dictate human-like action execution by influencing behavioral appropriateness}.
    Two cases to show the influence of the contextual situation for performing human-like actions. 
    Case I (``hugging friend'') requires an open-arm posture with elevated elbows to express affection, while Case II (``boxing with friend'') demands guarded elbows for defensive intent, despite nearly identical end-effector positions. 
    These full-body pose adaptations to contextual demands constitute \emph{behavioral appropriateness}, i.e., the alignment of motions with contextual situations.
    }
    \label{fig:Description}
\end{figure}

Humanoid robots play an pivotal role in human-robot interaction (HRI), where the behavioral human-likeness significantly influences user perception and acceptance~\cite{Planner}.
According to the uncanny valley theory, subtle deviations in robot behaviors that closely approximate human actions can elicit profound discomfort for humans.
This phenomenon reveals a fundamental challenge when generating human-like behaviors for humanoid robots: simultaneously ensuring behavioral appropriateness while preserving motion similarity.
Generally, similarity emphasizes the faithful reproduction of human kinematic and dynamic patterns, ensuring that the robot’s behavior physically resembles human motion.
In contrast, appropriateness emphasizes that robot behaviors must comply with situational demands and human cognitive expectations, thereby ensuring that the generated actions are contextually meaningful and socially acceptable within a given scenario. 
As the example shown in Fig~\ref{fig:Description}, these behaviors must not only satisfy kinematic goals, which are the targets of traditional behavior planning methods for behavioral similarity, but also align with human expectations within specific scenarios~\cite{Behavioral_appro1}, i.e., modifying the actions to adapt the contextual situation for achieving behavioral appropriateness.

Despite progress in behavioral similarity via various methods, e.g., motion retargeting (IK)~\cite{DexMimicGen,aisyah2025tantangan,zhang2022kinematic} and imitation learning \cite{jang2022bc,annabi2024unsupervised,he2024learning}, behavioral appropriateness remains under-explored in extensive discussions.
Moreover, existing humanoid motion generation methods to ensure behavior similarity suffer from several limitations.
First, most of the current datasets~\cite{plappert2016kit} exhibit insufficient description of expressiveness of human motion, relying on $6$D joint positions which lacks semantic annotations that bridge contextual situations to behavioral appropriateness.  
Moreover, several human demonstrated datasets just use short and simple text (e.g., 'running') to represent human motion sequence, which can not describe the entail and rich contextual situations. 
Second, the open-loop mechanism during pose generation and retargeting processes leads to the body structural mismatches in the two stages, which results in action human-likeness and semantics degradation.
For example,~\cite{cheng2024expressive} typically maps the generated pose to humanoids without considering the current robot state into generation module, thereby the generated pose neglecting the physical constraints of humanoids.
Sequently, resulting the retargeting problem. 
Third, cross-embodiment adaptation remains unaddressed, impeding deployment on heterogeneous robots~\cite{IK_robot} for different potential usage.

To address these challenges, we propose \textit{HuBE}, a bi-level closed-loop framework, that generates human-like behaviors achieving both behavioral similarity and appropriateness.
First, we propose the HPose dataset through open-sourced datasets, which incorporates contextual semantics of behavioral situations through fine-grained language annotations (e.g., “You hold the bottom of the cardboard box with both hands and struggle to raise it above your head.”) and leverage $6$D poses to preserve the behavioral human-likeness.
Second, we introduce a closed-loop mechanism through implicit skeletal parameter adaptation in \textit{HuBE}.
Therefore, it can fuse multi-modalities, i.e., robot state, behavioral goal, and contextual situation, to enable the end-to-end integration of motion generation and robot control and avoid the structure mismatching problem.
Furthermore, the introduction of contextual situation in robot action planning process guaranteed the behavioral appropriateness while keeping the behavioral similarity. 
Finally, we propose a bone scaling operation for data augmentation to simulate the morphological distribution of state-of-the-art commercial humanoids, addressing robust adaptation across heterogeneous platforms.
This enables the trained action generation model to generalize across diverse robotic kinematic parameters, laying the foundation for the millimeter-level cross-embodiment compatibility that distinguishes our framework with other works.
The contributions are summarized as follows:
\begin{itemize}
    \item We propose a new perspective on human-likeness, defined by behavioral similarity and appropriateness, supported by HPose dataset with fine-grained annotations bridging motion semantics for context-aware pose generation.
    \item We propose a bi-level closed-loop framework that enables humanoid robots to generate actions satisfying both behavioral similarity and appropriateness, while eliminating structural mismatches between the motion generation and execution modules.
    \item We introduce a bone scaling-based data augmentation strategy that provides millimeter-level cross-embodiment compatibility, enabling robust deployment across heterogeneous humanoid robots without additional shape adaptation.
\end{itemize}

%% file: Section/Related_work.tex
\section{Related works}
\label{sec:Related_work}
\begin{figure*}[ht]
    \centering
    \includegraphics[width=0.9\linewidth]{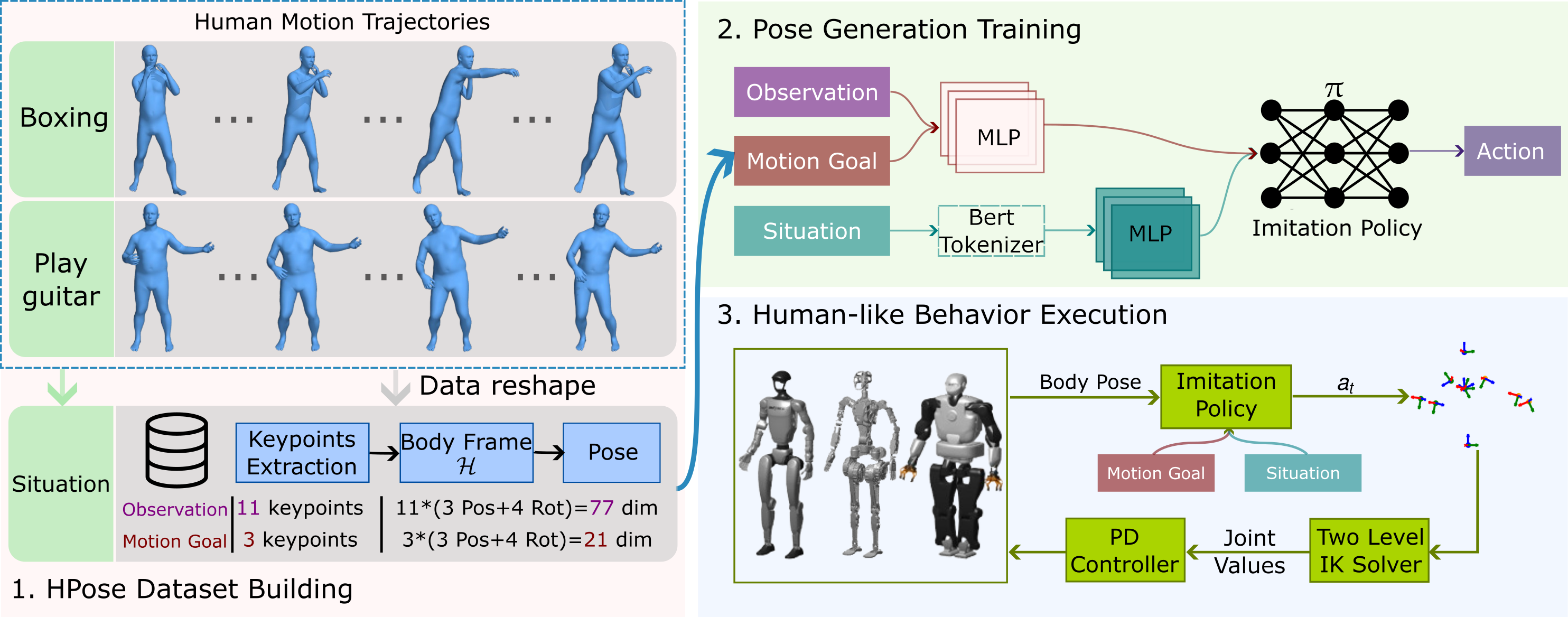}
    \caption{Overview of the whole algorithm. This algorithm includes three parts, i.e., building dataset, model training and algorithm implementation.}
    \label{fig:Module}
\end{figure*}

To ensure the behavioral similarity of humanoid robots, recent approaches focus on replicating human motion patterns through two paradigms: motion synthesis and imitation learning (IL). 
Motion synthesis methods generate human-like trajectories using auto-regressive models~\cite{HuMoR} or diffusion processes~\cite{Omnicontrol,VAE,MLD}, but often produce physically infeasible poses requiring post-processing. 
Moreover, a challenge for motion synthesis works~\cite{PirorMDM,Gmd} is that they cannot control well the added control signal for robot joints.
To tackle this issue, we concentrate on creating the robot's pose on a frame-by-frame basis using the robot state and joint goal, rather than generating an entire motion sequence.
IL methods~\cite{IL_whole_body,I-ctrl,Okami} learn control policies directly from human demonstrations, yet face challenges in handling heterogeneous robot morphologies and novel task constraints. 
Furthermore, some attempts~\cite{Open-television,Humanplus} ensure that humanoids learn autonomous skills from egocentric vision instead of third-view vision, which is closer to human behavioral habits.
However, this end-to-end learning method still suffers from the challenges of generalization performance for new tasks.
Consequently, we focus on learning a pose generation policy rather than a task skill to enhance the generalization ability of our model. 
Additionally, we introduce bone scaling operation strategies to boost the model's adaptability across different humanoid robots.

Research on behavioral appropriateness, rooted in social psychology and anthropology, examines the regulatory mechanisms through which social norms govern individual actions~\cite{Behavioral_appro2}. 
In HRI, empirical studies confirm that the congruence between robotic behaviors and user cognitive expectations directly dictates social acceptance for robots~\cite{Robot_behavioral_appro1,Robot_behavioral_appro2}.
Although existing efforts have achieved several accomplishments in behavioral appropriateness for mobile robots, such as verification models~\cite{Robot_behavioral_appro3}, it is still a challenge for humanoid robotics to perform contextually appropriate behaviors.
A critical limitation is the absence of systematic integration of contextual non-kinetic parameters (e.g., contextual situation) in recent human-like motion generation methods for humanoid robots, which severely constrains contextual appropriate motion synthesis. 
To resolve this, we propose a semantic-task fusion framework that synergizes situational semantics with robotic behavioral tasks, establishing a behavior generation paradigm compliant with human sociocognitive principles.

%% file: Section/Methodology.tex
\section{Methodology}
\label{sec:Methodology}

To drive humanoid robots to perform expressive human-like poses under the requirement of behavioral similarity and appropriateness, we propose the \textit{HuBE}, i.e., \textbf{hu}man-like \textbf{b}ehavior \textbf{e}xecution framework, which consists of three modules as shown in Fig.~\ref{fig:Module}, i.e., data processing, behavior generation, and behavior execution.  
The behavior generation module takes motion situations, current robotic observations, and target poses as inputs and generates human-like actions to reach the target. 
The behavior execution module then maps the generated human-like pose sequence to humanoid robots based on the physical characteristics provided in the robot pose data. 
\begin{figure}[t]
    \centering
    \includegraphics[width=\linewidth]{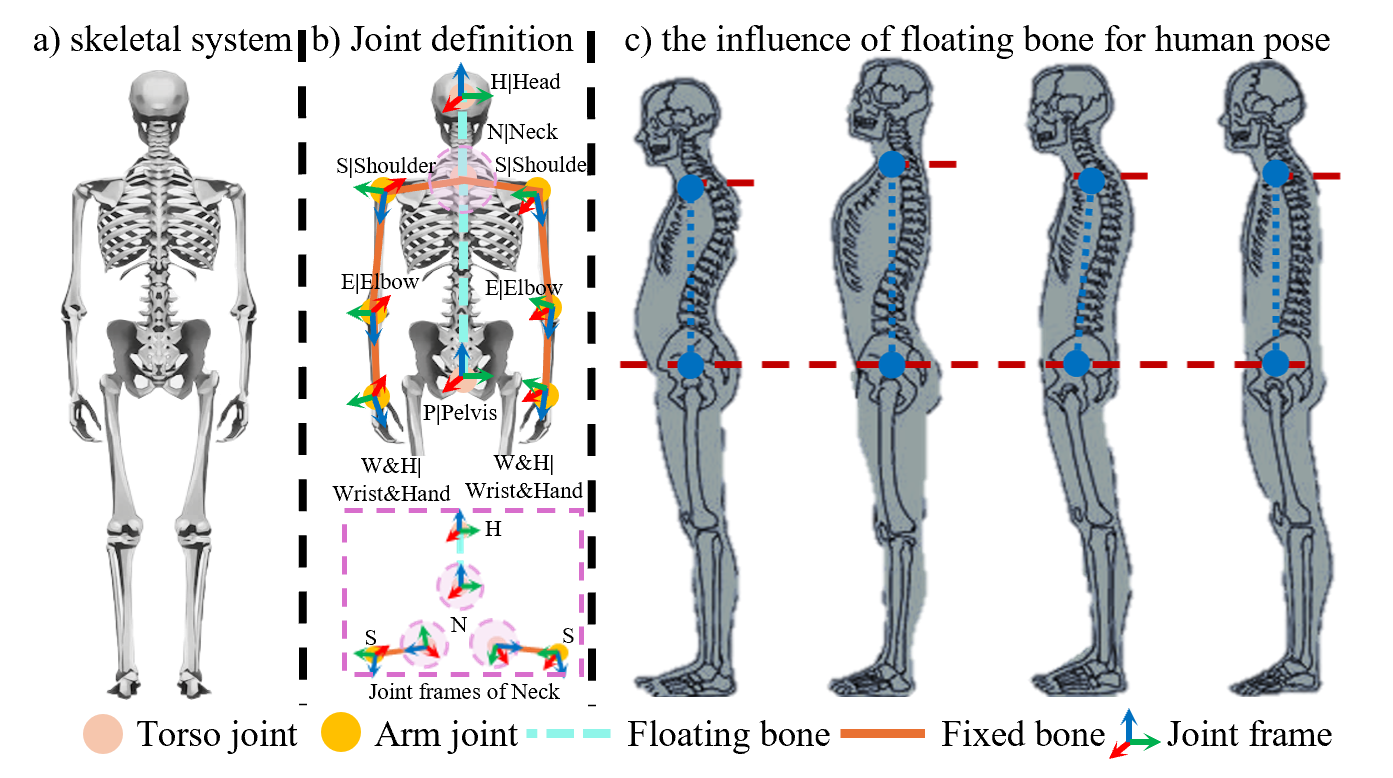}
    \caption{The human skeletal system introduction. 
    a) The composition of the human body's skeleton. 
    b) Simplified definition of human upper body chain $\mathcal{H}$ for expressive human pose dataset (HPose).  
    c) A simple example of the influence caused by floating bone.
    }
    \label{fig:bone_definition}
\end{figure}
\subsection{Data Specifications}
We enhanced a new dataset (HPose) in Table~\ref{tab:dataset} for this study by  open-sourced datasets, i.e.,  KIT \cite{plappert2016kit}, AMASS \cite{mahmood2019amass}, and Motion-X\cite{Motion-X} with three operations below:

\textbf{Data Definition.}
We reduced the body joints as depicted in Fig.~\ref{fig:bone_definition}, identifying $11$ main joints.
Specifically, the position of the left/right hand is identical to that of the left/right wrist.
Additionally, we introduced the rotational data that is represented by quaternions for each joint in our dataset. 
To enrich the contextual situation $l$ of human motions, we use the LLM (GPT-4o) to generate this description of given motion sequence with the annotation $\hat{l}$ in the original dataset.
In detail, we select two directed frames of one motion sequence with annotation $\hat{l}$ (i.e., a person is drinking water) and put them into LLM.
After reasoning the situation of these materials, the LLM generates a rich and detailed $l$ as \textit{"a human is drinking water while the cup is closely on the table in his left side. He is attempting to cause the cup with his right hand."}

 \textbf{Dataset Structure.}
We reformat motion data in each frame into the configuration $\{(s_i, g_i, l_i, a_i)\}$ to align with policy training needs.
Initially, we extract the 11 joint poses of the current frame, defining these as the apparent state $s_i$; 
Subsequently, we incorporate the poses of the three end-effectors, i.e., both hands and the head, from the subsequent frame as the goal state $g_i$. 
Ultimately, the 11 joint poses from the subsequent frame are utilized as the action $a_i$ to fulfill the goal $g_i$ based on the apparent state $s_i$. 
Additionally, a situational context $l_i$ is appended to characterize human behavior scenarios.

 \textbf{Data Augmentation.}
Due to the differences in body mechanisms between humans and humanoids, such as arm length, it is necessary to adjust the body shape of the model in the original dataset to generate poses for humanoids with various configurations.  
We get the augmented dataset $\mathcal{H}'$ (pseudo ground truth, \textbf{Pseudo GT}) through a bone scaling operation by updating the bone size with collected humanoid robot body parameters $R=\{r_i,r_k\}$ (same definition as Fig.~\ref{fig:bone_definition}) with Eq.~\ref{Eq:augment} and Algorithm~\ref{Algo:bone scaling}.
\begin{equation}
\mathcal{H}' = \bigcup_{(J_i, J_k) \in \mathcal{H}} \left( J_k' = J_i' + \frac{\overrightarrow{J_k - J_i}}{\| \overrightarrow{J_k - J_i} \|} \cdot \| \overrightarrow{r_k - r_i} \| \right)
\label{Eq:augment}
\end{equation}
Where the $\{(J_i,J_k)\}$ is the directed joint pair in the original dataset $\mathcal{H}$ (ground truth, \textbf{GT}), while the $\{(r_i,r_k)\}$ is the one in robot's body parameters $\mathcal{R}$.
Furthermore, $\mathcal{R}$ are collected from $9$ typical humanoid robots, such as Unitree H$1$ and PAL’s TALOS.

\begin{algorithm}[h]
  \KwData {Human body chain $\mathcal{H}=\{J_i\}$, robot body chain $\mathcal{R}=\{r_i\}$} 
  \KwResult {Augmented human body chain $\mathcal{H}'$}
  Generating directed joint pair set $D=\{(J_i,J_k)\}$ and joint frame set $\mathcal{H}' = \{J_0'\}$\;
  \While{$D\neq \emptyset$}
  {
      Randomly select $(J_i,J_k)$ \;
    \If{$(J_i,J_k) \in \mathcal{H}$ }
    {
    $V_k =\frac{J_k-J_i}{norm(J_k-J_i)}norm(r_k-r_i)$\;
    \If{$J_i' \in \mathcal{H}'$}
        {
        $J_k' = V_k + J_i’$\;
        $\mathcal{H}' = \mathcal{H}' \cup J_k'$\;
        $D = D-\{(J_i,J_k)\}$ \;
    }
    }
  }
  Return $\mathcal{H}'$
  \caption{Bone scaling for data augmentation}
  \label{Algo:bone scaling}
\end{algorithm}

To enable effective data augmentation for cross-embodiment compatibility, we simplify the skeletal system by categorizing bones into two types, i.e., fixed and floating bones, based on their size variability during human movement. 
As shown in Fig.~\ref{fig:bone_definition}-c, floating bones (e.g., the spine) exhibit length changes with body poses, while fixed bones remain stable; this distinction is critical for adapting human motions to humanoid robots via bone scaling operations. 

\begin{table}[h]\small
    \caption{Data source distribution of HPose dataset}
    \centering
    \renewcommand{\arraystretch}{1.2}
    \setlength{\tabcolsep}{9pt}
    \begin{tabular}{ccccc} \hline\rowcolor[HTML]{EFEFEF}
        Dataset & KIT  & AMASS & Motion-X & HPose \\ 
        \hline
        Motion  & 3912 & 5600  & 22413  & 31925 \\\rowcolor[HTML]{EFEFEF}
        Frame    & 2308K & 12457K  &  47339K & 62104K\\ 
        \hline
    \end{tabular}
    \label{tab:dataset}
\end{table}


\subsection{Pose Generation}
We aim to drive the humanoid robot to perform human-like poses based on a situational context $l$, current state $s$ and a motion target $g$. 
We formulate the humanoid robot motion generation problem as a Markov Decision Process \cite{sutton2018reinforcement} without a specific reward function \cite{torabi2018behavioral}, which is denoted by $(\mathcal{S}, \mathcal{A}, \mathcal{T}, \mathcal{G},l, \rho_0, \gamma)$. 
$\mathcal{S}$ is the state space containing all keypoint joint poses of the upper body of the robot, which implicitly reflects the body structure information, such as bone length.
$\mathcal{G}$ is the goal space that includes the target poses of end effectors, i.e., both hands and head.
$\mathcal{A}$ is the action space including all keypoint joint poses that the robot needs to take to reach the goal.
We use $\rho_0$ to denote the distribution of the initial state. 
$\mathcal{T}: \mathcal{S} \times \mathcal{A} \rightarrow \mathcal{S}$ is the transition function denoting the transition probability from the current state to the next state after taking an action. 
$\gamma$ is the discount factor.
The motion generator of the robot can be specified by a policy with parameters $\theta$, $\pi_{\theta}: \mathcal{S}, \mathcal{G}, l \rightarrow \mathcal{A}$, which specify the action that should be taken in the robot's current state (Eq.~\ref{Eq:policy}). 
The problem is to determine an imitation policy $\pi_{\theta}$, using a provided set of expert demonstrations $\{\xi_1, \xi_2, \cdots\}$. 
Therefore, we collect all the episodes in $\xi$ into a dataset $\{(s_i, g_i, l_i, a_i)\}$, and the objective is to maximize the probability of the demonstrated action, which can be formulated as follows,
\begin{equation}
    \theta^* = \arg\max_{\theta} \prod_{i=0}^N \pi_{\theta}(a_i | s_i, g_i, l_i),
    \label{Eq:policy}
\end{equation}
To achieve this imitation policy, two classical network architectures are used, i.e., MLP and transformer.

Considering the strategy of current motion generation methods which typically replicate fixed human motion sequences to achieve target poses, ignoring contextual situation constraints, several problems remain to be solved for our policy training process.
First, how to guarantee the behavioral appropriateness due to the exist of similar end-effector poses under varying situations.
Second, how to accurately generate the target behaviors which is not seen in human motion sequences (dataset).
To overcome these challenges, we propose \textbf{Input Alignment} and \textbf{Hindsight Training} strategies for stable human-like behaviors generation for humanoids.
Furthermore, the cross-platform compatibility of our policy is guaranteed by training with the augmented data.

 \textbf{Inputs Alignment.} To solve the first problem, we use $l$ to bridge the generated motion to appropriateness. 
By utilizing the $l$ of current motions, we first use the Bert tokenizer ($BERT(\cdot)$) to tokenize the $l_i$ into a fixed vector, then use an MLP ($MLP(\cdot)$) to map the humanoid proprioception ($s_i, g_i$) to the same size as the language vector, as shown in Fig.~\ref{fig:Module}-part 2 and Eq.~\ref{Eq:alignment}.
\begin{equation}
    v_{alig} = MLP(s_i \oplus g_i) \oplus MLP(BERT(l_i))
    \label{Eq:alignment}
\end{equation}

\textbf{Hindsight Training.} 
To generate an arbitrary motion pose for second problem, it is not efficient to directly select the next frame pose in the motion sequence as the target pose during the training.
Therefore, based on the idea of hindsight experience replay buffer \cite{andrychowicz2017hindsight}, which takes the state accomplished in the same trajectory $\tau$ as the final goal, we augment current datasets $\mathcal{D}$ by randomly taking the future poses in the frame window $H$ as the target pose we want the robot to achieve (shown in Eq.~\ref{Eq:Hindsight}). 
\begin{equation}
    \mathcal{D}_{\text{new}} = \bigcup_{\tau \in \mathcal{D}} \bigcup_{t=0}^{T} \left\{ \left( s_t,\ g_{t+k},\ l_t,\ a_{t+k} \right) \right\}^H_{k=1}
    \label{Eq:Hindsight}
\end{equation}


\subsection{Behavior Execution}
In this section, we aim to adopt the generated human-like actions into humanoid robots, i.e., converting actions from Cartesian space into the robot's joint space.
The challenge in this module is how to keep the human-likeness during the converting process, especially involving the movement of floating bones.  

\begin{figure}
    \centering
    \includegraphics[width=0.9\linewidth]{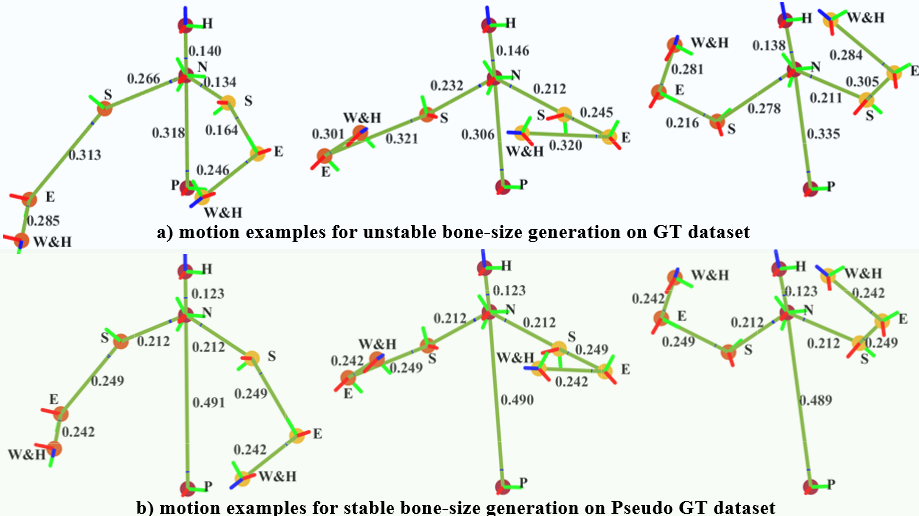}
    \caption{The generation results for bone-size stability. 
    The first row is the generated action based on the model trained on GT, while the second row is the results on the Pseudo GT. 
    The input body pose is captured from GR1, while the situational context is \textit{"Lift a heavy box over your head with both hands"}.}
    \label{fig:radar}
\end{figure}

To solve this problem, we propose a multi-constrained IK solver, i.e., a Closed-loop Inverse Kinematics (CLIK) algorithm based on the Pinocchio toolbox, to determine the robot joint control values corresponding to the target body pose.
Specifically, a two-step solver is implemented during the IK solving process. 
First, we calculate the robot torso joint angles $J_{\text{torso}}^*$, including joints of the neck and head ($\mathcal{E}_{\text{torso}}$). 
In this step, we emphasize the rotational pose of the target joint because the connected bone length in the torso (see Fig.~\ref{fig:bone_definition}-floating bone) varies with different body poses. 
Second, we create a reduced model by locking the torso joints using the Pinocchio toolbox and then calculating the dual-arm joint angles $J_{\text{arms}}^*$, such as wrists ($\mathcal{E}_{\text{arms}}$).
To enhance the stability of this IK solver, all the joint poses are processed by an SE(3) group filter, which is developed with Pinocchio’s SE(3) interpolation algorithm.
\begin{equation}
    \left\{
        \begin{aligned}
            J_{\text{torso}}^* = \arg\min_{J_{\text{torso}}} \sum_{i \in \mathcal{E}_{\text{torso}}} \left\| f_{\text{FK}}^{i}(J_{\text{torso}}) - p_{i}^{\text{target}} \right\|^2\\
            J_{\text{arms}}^* = \arg\min_{J_{\text{arms}}} \sum_{j \in \mathcal{E}_{\text{arms}}} \left\| f_{\text{FK}}^{j}(J_{\text{torso}}^*, J_{\text{arms}}) - p_{j}^{\text{target}} \right\|^2
        \end{aligned}
    \right.
    \label{Eq：IK_process}
\end{equation}

%% file: Section/Experiment.tex
\section{Results}
\label{sec:Experiment}
To validate the effectiveness of our methodology, corresponding evaluation metrics are provided. 
Specifically, two evaluation methods are proposed to assess the behavioral similarity. 
\begin{itemize}
    \item End-effector accuracy (E-A) is designed to measure the pose accuracy of the end-effectors, including both hands and head.
    \item Human similarity (H-S) aims to describe the pose consistency by considering all selected body markers.
\end{itemize}
For each evaluation method, we provide two metrics called Mean Per Joint Position Error (MPJPE)~\cite{siMLPe}  and Mean Per Joint Orientation Error (MPJOE) to measure the error of position and orientation, respectively.
To evaluate the behavioral appropriateness, three metrics are proposed as below:
\begin{itemize}
    \item Fréchet Motion Distance (FMD): measure the distributional similarity between generated motions and authentic human motions in the feature space similar as FID metric~\cite{motiondiffuse}. This metric works based on the premise that the human demonstrated motion in dataset is appropriate.
    \item Multimodal Distance (MM-Dist): calculate average distances between situation context and generated motions.
    \item R-Precision~\cite{motiondiffuse}: quantify the semantic discriminability of generated motions by establishing a text-to-motion cross-modal retrieval task.
\end{itemize}

\begin{table}
\centering
\renewcommand{\arraystretch}{1.2}
\setlength{\tabcolsep}{9pt}
\caption{Bone-size stability results [meter]}
\begin{tabular}{ccccc}
\hline \rowcolor[HTML]{EFEFEF}
Dataset             & Bone Type & GR1      & TALOS    & G1       \\ \hline
\multirow{2}{*}{GT} & Fixed     & 0.18     & 0.14     & 0.23     \\ 
                    & Floating  & 0.31     & 0.26     & 0.283    \\ \rowcolor[HTML]{EFEFEF}
GT \&               & Fixed     & \textbf{1.3e-4} & \textbf{2.4e-3} & \textbf{4.4e-4} \\ \rowcolor[HTML]{EFEFEF}
Pseudo GT           & Floating  & \textbf{4.1e-4} & \textbf{6.7e-3} & \textbf{6.8e-4} \\ \hline
\end{tabular}
\label{Table:implicit_results}
\end{table}


\begin{table*}[h]\small
\caption{The qualitative analysis results}
\renewcommand{\arraystretch}{1.2}
\setlength{\tabcolsep}{9pt}
\centering
\begin{tabular}{ccccccccc}
\hline
\rowcolor[HTML]{EFEFEF} 
\cellcolor[HTML]{EFEFEF}                        & \cellcolor[HTML]{EFEFEF}                            & \multicolumn{2}{c}{\cellcolor[HTML]{EFEFEF}MPJPE} & \multicolumn{2}{c}{\cellcolor[HTML]{EFEFEF}MPJOE} & \cellcolor[HTML]{EFEFEF}                      & \cellcolor[HTML]{EFEFEF}                      & \cellcolor[HTML]{EFEFEF}                              \\
\rowcolor[HTML]{EFEFEF} 
\multirow{-2}{*}{\cellcolor[HTML]{EFEFEF}Model} & \multirow{-2}{*}{\cellcolor[HTML]{EFEFEF}Frequency} & E-A$\downarrow$                     & H-S$\downarrow$                     & E-A$\downarrow$                     & H-S$\downarrow$                     & \multirow{-2}{*}{\cellcolor[HTML]{EFEFEF}FMD$\downarrow$} & \multirow{-2}{*}{\cellcolor[HTML]{EFEFEF}MM-Dist$\downarrow$} & \multirow{-2}{*}{\cellcolor[HTML]{EFEFEF}R-Precision $\uparrow$} \\ \hline
MLP(w)$^0$                                         & $240$                                                 & $0.135 $                  & $0.181$                   & $0.279$                   & $0.237$                    & $3.56$                                              & $10.4$                                              & $0.317$                                                      \\
\rowcolor[HTML]{EFEFEF} 
MLP(w)                                          & $60$                                                  & $0.091$                   & $0.127$                   & $0.219$                   & $0.178$                   & $3.02$                                              & $9.07$                                              & $0.341$                                                      \\
MLP(w)                                          & $15$                                                  & $0.056$                   & $0.077$                   & $0.172$                   & $0.131$                   & $1.95$                                              &   $6.74$                                            &  $0.384$                                                     \\
\rowcolor[HTML]{EFEFEF} 
MLP(w/o)$^1$                                        & Random                                              & $0.031$                   & $0.45$                   & $0.121$                   & $0.084$                   & $-$                                              &  $-$                                             & $-$                                                      \\
MLP(w)                                          & Random                                              & $0.015$                   & $0.026$                   & $0.044$                   & $0.043$                   &   $0.951$                                            &  $3.22$                                             & $0.492$                                                      \\
\rowcolor[HTML]{EFEFEF} 
Transformer(w/o)$^2$                                    & Random                                              & $0.023$                   & $0.031$                   & $0.071$                   & $0.058$                   &  $-$                                             &$-$                                               & $-$                                                      \\
Transformer(w)$^3$                                      & Random                                              & \textbf{0.008}                   & \textbf{0.019}                   & \textbf{0.039}                   & \textbf{0.036}                   &    \textbf{0.737}                                          & \textbf{3.07}                                            & \textbf{0.508}                                                     \\ 
\hline
\end{tabular}
\label{Table:Qualitative}
\end{table*}

\subsection{Multi-Embodiment Testing}
To evaluate the cross-embodiment capability of our works, two experiments are provided. 
Specifically, the bone-size stability is evaluated to determine if the generation module can generate actions satisfying skeletal systems for various humanoids.
Additionally, a retargeting accuracy testing is implemented to evaluate performance of execution module.

\textbf{Bone-size stability.}
Bone size is a fundamental characteristic of the human body and significantly influences the expressiveness of behaviors.
To verify its stability, we examine the generated results of two different bone types (fixed and floating) for three classical humanoid robots in various human motions.
The stability parameter is calculated by the average absolute error between the bone lengths of the input robot pose and the generated poses.

The results in Table~\ref{Table:implicit_results} show that the bone-size stability in the GT dataset is poor for both bone types of all the selected humanoid robots.
The reason is that the trained policy is not familiar with the input humanoid bone structure; 
thereby, the generated pose may be with wrong bone size (see Fig.~\ref{fig:radar}).
Although the motion retargeting strategy is widely used to adapt unmatched poses to humanoid robots, it is still faced with insurmountable defects, i.e., human-likeness loss during the motion retargeting processing as discussed in \cite{cheng2024expressive}.
To address this problem, we aim to produce human-like poses for humanoids while adhering to their skeletal system constrains.
Therefore, we add the augmented the pseudo GT to re-train the policy.
Fortunately, the generation error is optimized to the millimeter level, which means the bone-size stability for various humanoid robots is acceptable.
Additionally, in both datasets, the floating bone shows greater sensitivity compared to the fixed bone as shown in Table~\ref{Table:implicit_results} and Fig.~\ref{fig:radar}. 
This increased sensitivity arises because the floating bone's size fluctuates with different movements, even when the body configuration remains constant.

\begin{figure}[h]
    \centering
    \includegraphics[width=0.9\linewidth]{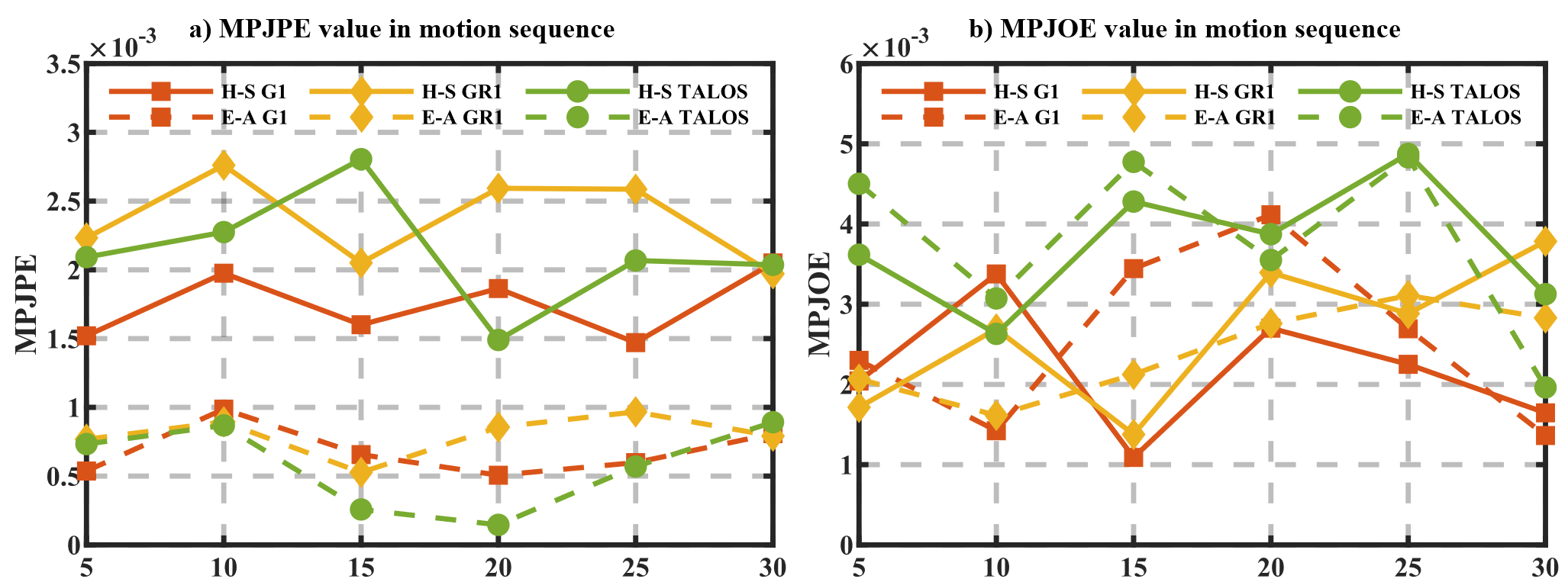}
    \caption{The retargeting accuracy between generated actions and the robot's actions for three humanoids in different frames. }
    \label{fig:Cross-platform}
\end{figure}

\begin{figure}[t]
    \centering
    \includegraphics[width=0.9\linewidth]{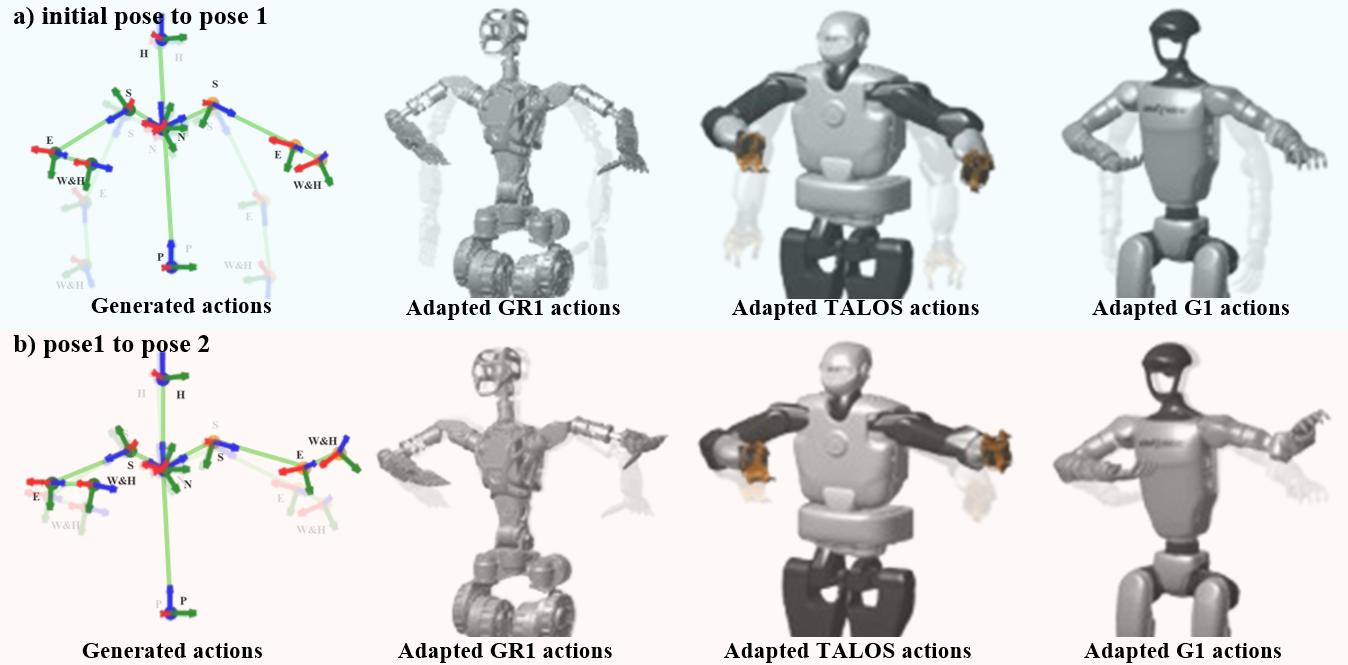}
    \caption{The implementation of our framework for expressive human-like behaviors in different humanoid robots. From left to right, the three popular humanoid robots are Fourier's GR1, Unitree's G1, and PAL's TALOS. }
    \label{fig:Multi-platform}
\end{figure}

  \textbf{Retargeting accuracy.}
We testing the retargeting accuracy by implementing it in various humanoids, e.g., GR1, with generated motion in the situational context \textit{"a person is start to dancing with partner."}.
As indicated in Fig.~\ref{fig:Cross-platform}, the quantitative results proved that our framework can be adopt to the selected humanoid robots while keeping action accuracy and human similarity with small value of MPJPE and MPJOE for both E-A and H-S.
However, the MPJPE value of H-S is bigger than that of E-A as we emphasize the rotational pose in pose retargeting process for floating bones due to its unique character.
Fortunately, the accuracy of the retargeting module is guaranteed by keep small$\&$similar MPJOE and acceptable MPJPE in both E-A and H-S.
A qualitative demonstration is presented in Fig.~\ref{fig:Multi-platform}, where two continuing frames in a planned human-like motion with given situation are selected.
For each motion frame in different color boxes, the pose with light color is the initial state while the pose with dark color is the planned pose based on given conditions, i.g., goals.
It is proved that our framework has good adaptability for action planning and driving most humanoid robots.

\subsection{Ablation Study}
We implement our framework using two classical network architectures, MLP-based and Transformer-based policies, to assess its performance. 
The KIT\_ML dataset was selected for this evaluation due to its consistent data frequency of $240$ Hz.

  \textbf{Ablation study of frequency.} 
To evaluate the influence of data frequency on our framework performance, we train our network with four types of frequency, i.e., $240$Hz (original), $60$Hz, $15$Hz and random frequency.
In the testing process, a target pose was randomly sampled in the test dataset. 
The testing results in Table~\ref{Table:Qualitative} indicate that the random sampling method (hindsight training) achieved the best values of all evaluation metrics for behavioral similarity and appropriateness compared to others with fixed frequency.
The reason is that the two training frames for fixed frequency groups are too close to identify the difference between the frames when the dataset frequency is too high.
The inference is supported by the tendency of experimental results for all the evaluation metrics while the dataset frequency is decreased.
As the motion speed of demonstrated human action is not uniformly distributed in the time domain, it still cannot achieve optimal framework performance even with small recording frequencies, such as $15$Hz.
Due to the random frequency strategy enlarging the dataset and enriching the data range by randomly selecting goals from the trajectory, the performance of E-A and H-S is improved.
Furthermore, the hindsight training method also enriched the motion semantics of human-demonstrated behavior.
Therefore, the contextual appropriateness is enhanced, which is indicated by the metrics of FMD, MM-Dist, and R-Precision.

\textbf{Ablation study of motion situation.}
To explore the effects of the motion situation $l$ for human-like motion generation, an ablation study with two architectures is carried out, and the results are shown in Table~\ref{Table:Qualitative}.
Flag (w) means the model is trained by the dataset with motion situation $l$. Otherwise, it is marked as (w/o).
The results show that the model performance for both behavioral similarity and appropriateness trained with motion situation $l$ is better than that without motion situation.
The reason for this promotion is that the appropriate human action pose is related to the contextual situation and motion target, such as \textit{"lifting a box while avoiding collisions with the table on your left"}.
Unfortunately, this  information of contextual situation cannot be explicitly represented by human actions data.
Thereby, the dataset with situation context $l$ is more essential for the human-like behavior planning task for humanoid robots.

\begin{figure}[t]
    \centering
    \includegraphics[width=0.9\linewidth]{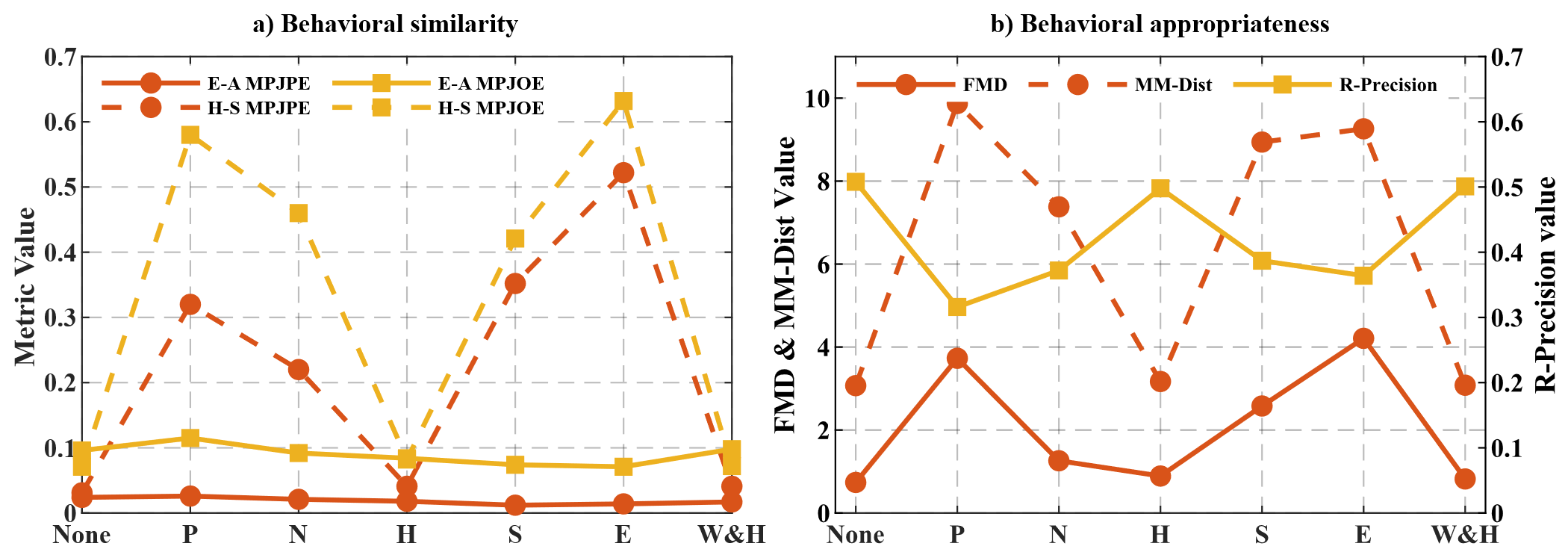}
    \caption{The results of ablation for observation input in our framework. The X-axis means the locking data, e.g., the "None" means that we provide whole observation, while "N" means the Neck data in observation is missing.}
    \label{fig:Ablation_observation}
\end{figure}

  \textbf{Ablation study of observation.}
The results in Fig.~\ref{fig:Ablation_observation} indicate that the missing observation of "pelvis, neck, shoulder, and elbow" influence E-A in both MPJPE and MPJOP is less.
However, it is hugely damaged the H-S and behavioral appropriateness.
The reason is that lack of observation of these joints would destroy the body chain which will suffer the generation performance.
In contract, the missing observation of "Hand and Wrist$\&$hand" has less impact on both behavior similarity and appropriateness, due to the end-effectors' human-like poses along with the part of our training target, i.e., generating a pose to meeting the goals.
Therefore, it is necessary to ensure the integrity of the observations.

\subsection{Comparative Results}
To prove the efficiency of our method, we compare our framework with other state-of-the-art works.
Three classical methods are selected in the comparison study,
i.e., IK algorithm, siMLPe~\cite{siMLPe}, and HumanPlus~\cite{Humanplus}.

  \textbf{Methods comparison.}
For behavioral similarity, the results shown in Table~\ref{Table:Comperation} demonstrated that our approach achieves better results on both the MPJOE and MPJPE metrics compared to the siMLPe and HumanPlus models. 
This outcome arises because the selected model just considers the behavioral traits in human-demonstrated sequential motion, which are also widely accepted by motion synthesis. 
Consequently, these methods do not perform as expected when dealing with unseen motion traits in training dataset, which are common in daily tasks.
Fortunately, our approach has the capability to address this challenge due to its innovative mechanism.
By comparing with IK, our framework gets poorer but acceptable results in E-A with both MPJRE and MPJOE metrics as the E-A is the only target to optimize for the IK solver.
Specially, the error of E-A in our method is mainly introduced in generation process.
However, we can achieve better performance in H-S than IK, which ensures that the generated pose for humanoid robots is more similar to humans.
Another advantage of our method ($0.0207$s) is time-saving when compared with IK ($0.3618$s), which is based on a searching mechanism.
Therefore, our method is a good choice for human-like pose generation tasks. 
For behavioral appropriateness, our method can achieve smaller FMD and MM-Dist values, which means that our method can generate higher quality and closer distribution of real actions than the selected method.
Furthermore, the higher the R-Precision value indicates that our method can get better semantic matching accuracy between the generated action and the situation description.
Therefore, the behavioral appropriateness of the generated action in our module is guaranteed.

\begin{figure*}
    \centering
    \includegraphics[width=0.9\linewidth]{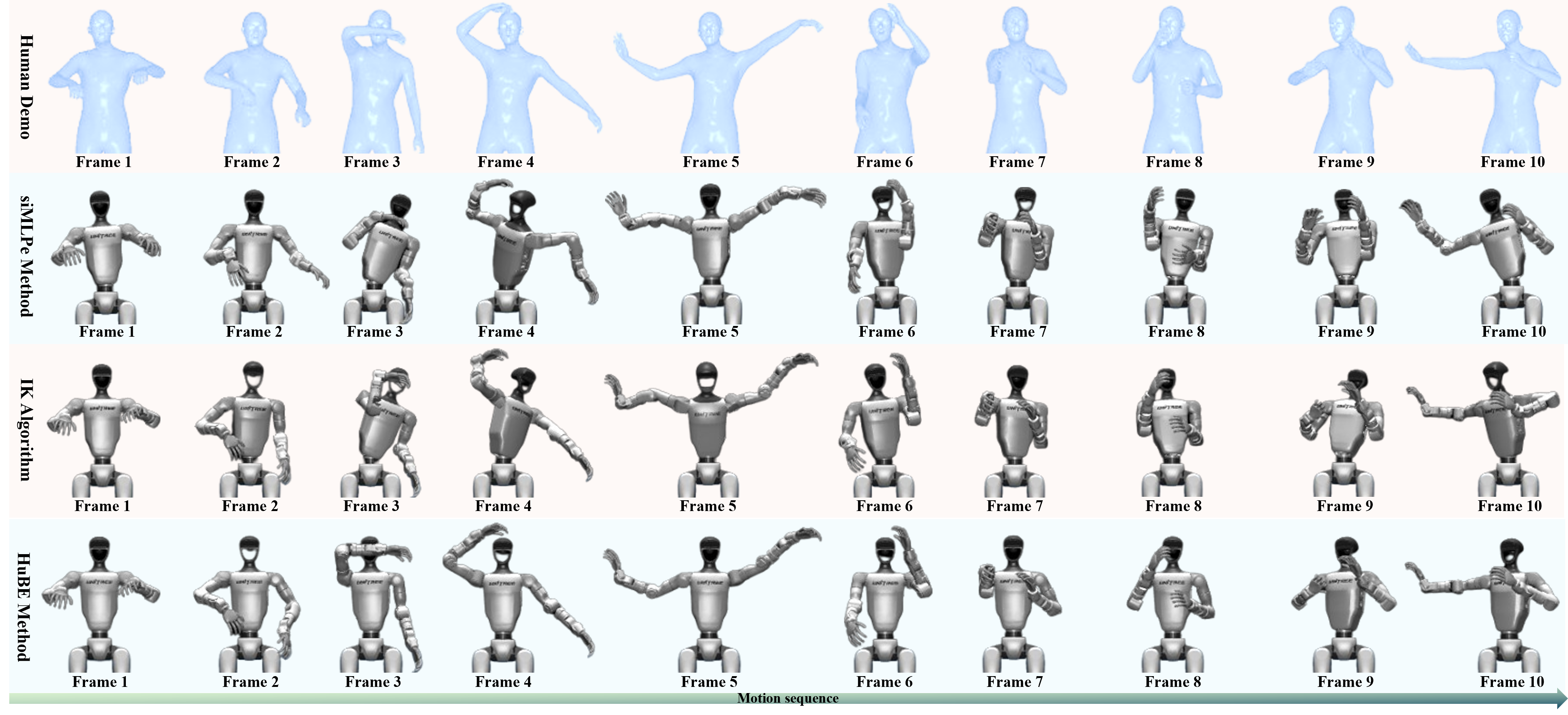}
    \caption{
    An example of action-wise results (“Playing Kung Fu”) performed by the humanoid robot G1. 
    Motion sequences are shown from left to right.
    From top to bottom: (1) human demonstration data from our dataset, 
    (2) results generated by siMLPe, 
    (3) motions planned by the IK algorithm with target pose $g$, and (4) results produced by our proposed method (HuBE) conditioned on $g$ and contextual situation $l$.
    Due to the physical constraints of G1, the target $g$ refers to the wrist poses of both arms.
    }
    \label{fig:Real_robot_action}
\end{figure*}

  \textbf{Action-wise.}
We also evaluated our method with different actions, and the action-wise results are also shown in Fig.~\ref{fig:Real_robot_action} and Table~\ref{Table:Comperation}.
By analyzing the results, we know that our framework is well adapted to various human actions as the value of evaluation metrics for different actions is stable and small. 
Consequently, our framework can effectively generate stable human-like poses for humanoids over various tasks when compared with other works.

\begin{table*}[h]\small
\centering
\caption{The action-wise results for different methods}
\renewcommand{\arraystretch}{1.2}
\setlength{\tabcolsep}{9pt}
\begin{tabular}{ccccccccc}
\hline
\rowcolor[HTML]{EFEFEF} 
\cellcolor[HTML]{EFEFEF}                           & \cellcolor[HTML]{EFEFEF}                        & \multicolumn{2}{c}{\cellcolor[HTML]{EFEFEF}MPJPE} & \multicolumn{2}{c}{\cellcolor[HTML]{EFEFEF}MPJOE} & \cellcolor[HTML]{EFEFEF}                      & \cellcolor[HTML]{EFEFEF}                      & \cellcolor[HTML]{EFEFEF}                              \\
\rowcolor[HTML]{EFEFEF} 
\multirow{-2}{*}{\cellcolor[HTML]{EFEFEF}Action}   & \multirow{-2}{*}{\cellcolor[HTML]{EFEFEF}Model} & E-A                      & H-S$\downarrow$                    & E-A                       & H-S$\downarrow$                   & \multirow{-2}{*}{\cellcolor[HTML]{EFEFEF}FMD$\downarrow$} & \multirow{-2}{*}{\cellcolor[HTML]{EFEFEF}MM-Dist$\downarrow$} & \multirow{-2}{*}{\cellcolor[HTML]{EFEFEF}R-Precision$\uparrow$} \\ \hline
& IK                                              & $0.00832$                  & $0.324$                  & $0.000853$                  & $1.62$                  &   $-$                                            & $-$                                              &  $-$                                                     \\

& HumanPlus                                       &  $0.109$                        &  $0.184$                      & $1.73$                           & $3.18$                      & $-$                                              &  $-$                                             & $-$                                                      \\
& siMLPe                                          & $0.0841$                   & $0.138$                  & $1.42$                      & $3.21$                  &    $1.73$                                           & $6.25$                                              &  $0.361$                                                     \\
\multirow{-5}{*}{Dancing}                          
& Ous                                             & \textbf{0.0196}                  & \textbf{0.0337}                  & \textbf{0.0743}                     & \textbf{0.0543}                 & \textbf{0.749}                                              &  \textbf{3.71}                                             &  \textbf{0.484}                                                     \\
\rowcolor[HTML]{EFEFEF} 
\cellcolor[HTML]{EFEFEF}                           
& IK                                              & $0.00785$                  & $0.425$                  & $0.000742$                  & $2.24$                  &   $-$                                            &$-$                                               & $-$                                                      \\
\rowcolor[HTML]{EFEFEF} 
\cellcolor[HTML]{EFEFEF}                                                    
& HumanPlus                                       & $0.139$                         & $0.195$                       & $2.52$                          &  $2.65$                     &$-$                                               & $-$                                              &$-$                                                       \\
\rowcolor[HTML]{EFEFEF} 
\cellcolor[HTML]{EFEFEF}                           
& siMLPe                                          & $0.0728$                   & $0.114$                  & $2.12$                      & $2.64$                  & $1.68$                                              &  $6.82$                                             &   $0.337$                                                    \\
\rowcolor[HTML]{EFEFEF} 
\multirow{-5}{*}{\cellcolor[HTML]{EFEFEF}Exercise} 
& Ous                                             & \textbf{0.0176}                   & \textbf{0.0297}                 & \textbf{0.0476}                     & \textbf{0.0403}                 &  \textbf{0.763}                                             &  \textbf{3.28}                                             &   \textbf{0.492}                                                    \\
& IK                                              & $0.00983$                  & $0.524$                  & $0.000947$                  & $1.93$                  & $-$                                              & $-$                                              &  $-$                                                     \\
& HumanPlus                                       & $0.0972$                         & $0.115$                       & $2.48$                          &  $294$                     &  $-$                                             & $-$                                              & $-$                                                      \\
& siMLPe                                          & $0.0736$                   & $0.101$                  & $2.62$                      & $3.37$                  &   $1.94$                                            &  $7.28$                                             &  $0.317$                                                     \\
\multirow{-5}{*}{Playing}                         
& Ous                                             & \textbf{0.0188}                   & \textbf{0.0299}                 & \textbf{0.0553}                     & \textbf{0.0427}                 & \textbf{0.823}                                              & \textbf{4.01}                                              &  \textbf{0.466}                                                     \\
\rowcolor[HTML]{EFEFEF} 
\cellcolor[HTML]{EFEFEF}                           
& IK                                              & $0.00941$                  & $0.183$                  & $0.000318$                  & $2.12$                  &$-$                                               & $-$                                              & $-$                                                      \\
\rowcolor[HTML]{EFEFEF} 
\cellcolor[HTML]{EFEFEF}                                                     
& HumanPlus                                       &   $0.0862$                       & $0.103$                       &$1.80$                           &  $1.75$                     & $-$                                              &$-$                                               &$-$                                                       \\
\rowcolor[HTML]{EFEFEF} 
\cellcolor[HTML]{EFEFEF}                           
& siMLPe                                          & $0.0761$                   & $0.129$                  & $1.65$                      & $1.96$                  &   $1.69$                                            & $6.04$                                              &  $0.493$                                                     \\
\rowcolor[HTML]{EFEFEF} 
\multirow{-5}{*}{\cellcolor[HTML]{EFEFEF}Running}  
& Ous                                             & \textbf{0.0197}                   & \textbf{0.0295}                 & \textbf{0.0441}                     & \textbf{0.0403}                 &  \textbf{0.715}                                             & \textbf{3.27}                                              & \textbf{0.531}                                                      \\
\hline
\end{tabular}
\label{Table:Comperation}
\end{table*}

%% file: Section/Conclusion.tex
\section{Conclusion}
\label{sec:Conclusion}
In this paper, we presented \textit{HuBE}, a bi-level closed-loop framework for human-like behavior execution in humanoid robots. 
By integrating robot state, goal poses, and contextual situations, our framework enables the generation of behaviors that satisfy both behavioral similarity and appropriateness. 
To support this framework, we developed HPose, a context-enriched dataset with $6$D joint pose representation and situational annotations, and introduced a bone scaling-based data augmentation strategy to ensure millimeter-level cross-embodiment compatibility across heterogeneous humanoid robots.
Extensive experiments on multiple commercial platforms demonstrated that \textit{HuBE} significantly outperforms state-of-the-art methods. 
These results highlight the potential of \textit{HuBE} to serve as a transferable foundation for scalable and socially acceptable human-like behavior execution.
One constraint of \textit{HuBE} is that our planner primarily emphasizes the human-like movements of the robot's upper body. 
However, the movements of the lower body are also significant in influencing human actions. 
Therefore, our upcoming research will focus increasingly on designing full-body expressive behaviors that are feasible and mimic human movements.